\documentclass{article}

\PassOptionsToPackage{numbers, sort&compress}{natbib}

\usepackage[preprint]{neurips_2021}

\usepackage[utf8]{inputenc} 
\usepackage[T1]{fontenc}    
\usepackage{hyperref}       
\usepackage{url}            
\usepackage{booktabs}       
\usepackage{amsfonts}       
\usepackage{nicefrac}       
\usepackage{microtype}      
\usepackage{xcolor}         

\usepackage{amsmath}
\usepackage{amsthm}
\usepackage{graphicx}
\usepackage[ruled, linesnumbered]{algorithm2e}

\newtheorem{definition}{Definition}
\newtheorem{theorem}{Theorem}
\newtheorem{corollary}{Corollary}

\title{Graph Inference Representation: Learning Graph Positional Embeddings with Anchor
  Path Encoding}

\author{%
  Yuheng Lu, Jinpeng Chen \\
  School of Computer Science (National Pilot School of Software)\\
  Beijing University of Posts and Telecommunications \\
  Beijing, China \\
  \texttt{yuheng.lu@bupt.edu.cn}, \texttt{jpchen@bupt.edu.cn}\\
  \And
  Chuxiong Sun, Jie Hu \\
  Emerging Technology Research Division \\
  China Telecom Research Institute \\
  Beijing, China \\
  \texttt{chuxiongsun@gmail.com}, \texttt{hujie1@chinatelecom.cn} \\
}

\begin{document}

\maketitle

\begin{abstract}
  Learning node representations that incorporate information from graph structure
  benefits wide range of tasks on graph.
  The majority of existing graph neural networks (GNNs) have limited power in
  capturing position information for a given node.
  The idea of positioning nodes with selected anchors has been exploited,
  yet mainly relying on explicit labeling of distance information.
  Here we propose Graph Inference Representation (GIR),
  an anchor based GNN model encoding path information related to pre-selected anchors for each
  node.
  Abilities to get position-aware embeddings are theoretically and experimentally
  investigated on GIR and its core variants.
  Further, the complementarity between GIRs and typical GNNs
  is demonstrated.
  We show that GIRs get outperformed results in position-aware scenarios,
  and performances on typical GNNs could be improved by fusing GIR embeddings.
\end{abstract}

\section{Introduction}\label{section:introduction}

Graph, as an important data structure, is a powerful tool to represent
ubiquitous relationships in the real world.
Learning vector representations for graph data, benefits many downstream tasks on
graph such as node classification \cite{kipf2017semi} and link prediction
\cite{Zhang2018}.
Many graph representation learning methods have been proposed recently,
among those, Graph Neural Networks (GNNs), inheriting the merits of neural
networks, have shown superior performance and become a much popular choice.

\begin{figure}
  \centering
  \includegraphics[width=0.75\linewidth]{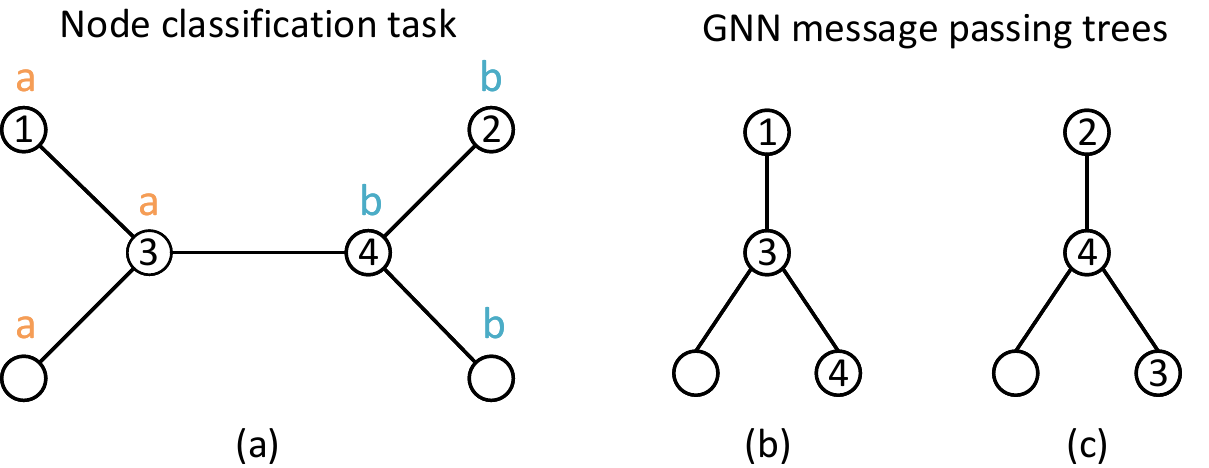}
  \caption{Node classification on the structral symmetric cases.
    \textbf{(a)}
    Example unattributed graph that assign different labels (a and b) on nodes with
    structral symmetry.
    \textbf{(b\&c)}
    MPNNs message passing tree of node $v_1$ and $v_2$, message propagate from leaves to
    the root.
    \textbf{Distinguishability:}
    For $v_1$ and $v_2$,
    (1)No anchors: indistinguishable due to identity sub-structure;
    (2)$v_3$ or $v_4$ as anchor: distinguishable by different position from the
    anchor node;
    (3)$v_3$ and $v_4$ as anchors: distinguishable if the order of $v_3$ and $v_4$
    is aware, indistinguishable otherwise.}
  \label{fig-pgnn}
\end{figure}

Existing GNN models mainly follow the message passing neural network (MPNN)
\cite{gilmer2017neural} pattern,
which stacks MPNN layers that
aggregate information from neighborhoods and then update representations for each node.
Typical MPNNs have been shown to lack of ability to capture the
position information within graph \cite{you2019position},
without distinguishable node/edge attributes, nodes in different part of the graph
with topologically equivalent neighborhood structures will be embedded into the same 
representation by typical MPNNs alone \cite{you2019position,
  li2020distance}, Figure \ref{fig-pgnn} gives an example.

Researchers have developed methods to alleviate this issue.
Some earlier works adopt one-hot encodings as extended node attributes
\cite{kipf2017semi}.
More recent anchor based methods \cite{you2019position, liu2019gnn} select anchor nodes as positioning base,
use position information related to anchors to break the structural symmetry,
Figure \ref{fig-pgnn} shows some cases of this strategy.
Existing anchor based methods mainly directly pass message from anchors to each node,
for modeling positional relationship,
distance information from anchors are explicitly assigned as node attributes.
However, this strategy ignores the graph structure information to some extent, 
and the performance is strongly relying on the quality of the explicitly assigned information.

In this paper, we propose an anchor based GNN model,
termed Graph Inference Representation (GIR),
aiming at encoding path information related to pre-selected anchor nodes.
GIRs first select fixed anchor nodes heuristically,
messages are propagated from anchors along paths to each node,
and outputs of the k-th layer encode k-order representations related to anchors.

Our contributions are summarized as follows:

\begin{enumerate}
\item We propose the Graph Inference Representation (GIR) model that
  follows the anchor based graph neural network (A-GNN) pattern.
  GIRs perform anchor path encoding by propagating message along paths from
  anchors.
\item We theoretically investigate the application of GIRs in position-aware
  scenarios,
  several variants are proposed to get more accurate positioning.
  We evaluate the performance of GIR and those variants for tasks in position-aware
  datasets,
  experimental results show that our position-aware GIRs archive generally
  higher performance.
\item We exploit the complementarity between GIRs and MPNNs,
  and adopt multi-modal fusion techniques to get benefit from this merit,
  experimental results demonstrate the improvement.
\end{enumerate}

\section{Related works}\label{section:related-works}

\subsection{Graph Neural Networks and anchor based methods}

A large amount of existing GNN models follow the MPNN pattern,
with different neighborhood aggregation and node representation update functions.
Graph Convolution Networks (GCN) \cite{kipf2017semi} uses normalized mean aggregator;
GraphSAGE \cite{hamilton2017inductive} concatenates node's representation and neighborhood information from
mean/max/LSTM aggregator;
Graph Attention Network (GAT) \cite{petar2018graph} introduces attention mechanism in the neighborhood
aggregation step.
Our GIR model uses the same aggregate and update propagation strategy as MPNNs,
but with different message propagation paths.

Anchor based GNN (A-GNN) methods follow the intuition that some nodes play more important role
in the graph.
Existing A-GNN models mainly follow the two stage pattern:
select anchor nodes first,
and then encode the information related to anchors.
Position-aware GNN (PGNN) \cite{you2019position} proposes to
position nodes with selected anchor nodes.
PGNNs select anchor node sets
randomly before running every forward of the model
to get a low distortion embedding with global position information,
an order-aware output layer was proposed to generate position-aware
embedding for each node from anchor node sets,
the output layer generates a vector representation whose elements
are computed by messages directly received from corresponding anchor node set.
The random anchor selecting strategy of PGNNs leads to unstable limitation,
some more recent work use fixed anchor nodes instead to overcome
this issue.
AGNN \cite{liu2019gnn} pre-selects fixed anchors by minimum point cover nodes algorithm;
GraphReach \cite{Nishad2020} follows the fix anchor setting,
and captures global positions of nodes by random walk reachability estimations
instead of the shortest path distance.
Other than selecting anchors as positioning base,
Graph Inference Learning (GIL) \cite{Xu2020Graph} views nodes in the training set as
anchors,
and boosts the semi-supervised node classification performance by
learning the inference of node labels on graph topology.
Our GIR model is designed for position-aware scenarios,
and differs existing A-GNN models in the anchor information encoding
strategy.

\subsection{Multi-modal Fusion}

Multi-modal fusion aims at integrating information from multiple modalities,
which could be categorized as early and late approaches \cite{baltruvsaitis2018multimodal}.
Early fusion, or feature fusion, makes decision on fused representation.
Representative methods of early fusion include gated fusion \cite{arevalo2017gated},
and bilinear model \cite{2017Multi, 2018Efficient}.
Late fusion, or decision fusion, make decisions on each uni-modal and fuses them
for the final decision.
Mixture of Experts (MoE) \cite{jacobs1991adaptive} is a representative late fusion model,
which incorporate a gating network to decide which expert to use for each input.
Widely used gating strategy in multi-modal fusion may face with fusion weights
inconsistency issue,
\cite{shim2019robust} suggest using uni-modal outputs loss to regularize fusion
weights.
We adapt MoE model for graph data,
and use it to exploit the complementarity between GIRs nad MPNNs,
strategies for fusion inconsistency issue are adopted.

\section{Preliminaries}\label{section:preliminaries}

\begin{table}
  \caption{Notations}
  \label{table-notation}
  \centering
  \begin{tabular}{cl}
    \toprule
    Notation & Description \\
    \midrule
    $\mathcal{G}$ & the input graph \\
    $\mathcal{V,E}$ & the node/edge set of $\mathcal{G}$ \\
    $v_i$ & the i-th node in $\mathcal{G}$ \\
    $\mathcal{X}$ & the node attributes of $\mathcal{G}$ \\
    $\mathcal{Z}$ & the node representations of $\mathcal{G}$ \\
    $x_i,z_i$ & the attribute/representation of $v_i$ \\
    $\mathcal{N}(v)$ & in-neighborhoods of node v in $\mathcal{G}$ \\
    $\mathcal{A}$ & the anchor node set \\
    $\mathnormal{A}$ & set of subsets of $\mathcal{A}$ \\
    $n$ & the number of nodes \\
    \bottomrule
  \end{tabular}
\end{table}

\subsection{Notations and problem definition}

A graph can be represented as $\mathcal{G=(V,{E})}$,
where $\mathcal{V}=\{v_1,\cdots ,v_n\}$ is the node set
and $\mathcal{E}=\{\langle v_i,v_j \rangle|v_i, v_j \in \mathcal{V}\}$ is the edge set.
Nodes are augmented with the feature set $\mathcal{X}=\{x_1,\cdots x_n\}$,
which is either input attributes or placeholders if no attributes available.
In-neighborhoods of node $v$ are represented as $\mathcal{N}(v)$.
Notations are summarized in Table~\ref{table-notation}

The GIR model is designed for node representation learning task,
which takes graph $\mathcal{G=(V,E)}$ with node attributes $\mathcal{X}$ as input, 
and embed nodes into d-dimensional vectors,
represented as $\mathcal{Z}=\{z_1,\cdots z_n\},z_i\in \mathbb{R}^d$.
The node representations are normally used in downstream tasks like node
classification and link prediction.

\subsection{Position-aware embeddings}

One goal of the anchor based GNN model is to utilize anchors as bases to encode
position aware information for each node.
To capture this intuition,
PGNNs \cite{you2019position} view embeddings as position-aware if shortest path distance
between node pairs could be reconstructed from their embeddings (Definition \ref{def-pos}).

\begin{definition}[Position-aware Embeddings]
  \label{def-pos}
  A node embedding $z_i=f_p(v_i),\forall v_i\in \mathcal(V)$ is position-aware
  if there exists a function $g(\cdot,\cdot)$ such that
  $d_{sp}(v_i,v_j)=g(z_i,z_j)$,
  where $d_{sp}(\cdot,\cdot)$ is the shortest path distance in $\mathcal{G}$.
\end{definition}

For models with fixed anchors,
the shortest path distance between two nodes may not path through anchors,
thus hard to satisfy definition \ref{def-pos}.
To capture the characteristic of position modeling for fixed anchors models,
here we define position-aware embeddings related to anchors (Definition
\ref{def-apos}).
Considering the differences of anchor utilization strategies (Figure
\ref{fig-pgnn}, b\&c),
our definition of position aware embeddings is made on a set of anchor node sets.

\begin{definition}[$\mathnormal{A}$-position-aware Embeddings]
  \label{def-apos}
  For $\mathnormal{A}\subset \mathcal{P(A)}, \bigcup_{\mathnormal{A}}=\mathcal{A}$, where $\mathcal{P(A)}$ is the
  power set of node set $\mathcal{A}$,
  a node embedding $z_i=f_p(v_i),\forall v_i\in \mathcal{V}$ is
  $\mathnormal{A}$-position-aware
  if there exists functions $\{g_{\mathcal{A}'}(\cdot)|\forall \mathcal{A}' \in \mathnormal{A}\}$ such
  that
  $g_a(z_i)=d_{sp}(v_i, \mathcal{A}')$
  where $d_{sp}(v, \mathcal{A}')$ is the shortest path distance between node v and
  node set $\mathcal{A}'$.
  If all elements in $\mathnormal{A}$ are singleton,
  brakets of sets inner $\mathnormal{A}$ could be omitted.
\end{definition}

\section{Proposed methods}\label{section:proposed-methods}

In this section, we illustrate the key aspects of GIR model.
We first describe the main framework and key components of GIRs.
Next we propose position-aware variants which enable more accurate positioning,
design choices and theoretical analysis are discussed.
Then, we describe the fusion strategy for exploiting the complementarity between GIRs
and MPNNs.

\subsection{The framework of GIRs}

\begin{figure}
  \centering
  \includegraphics[width=0.8\linewidth]{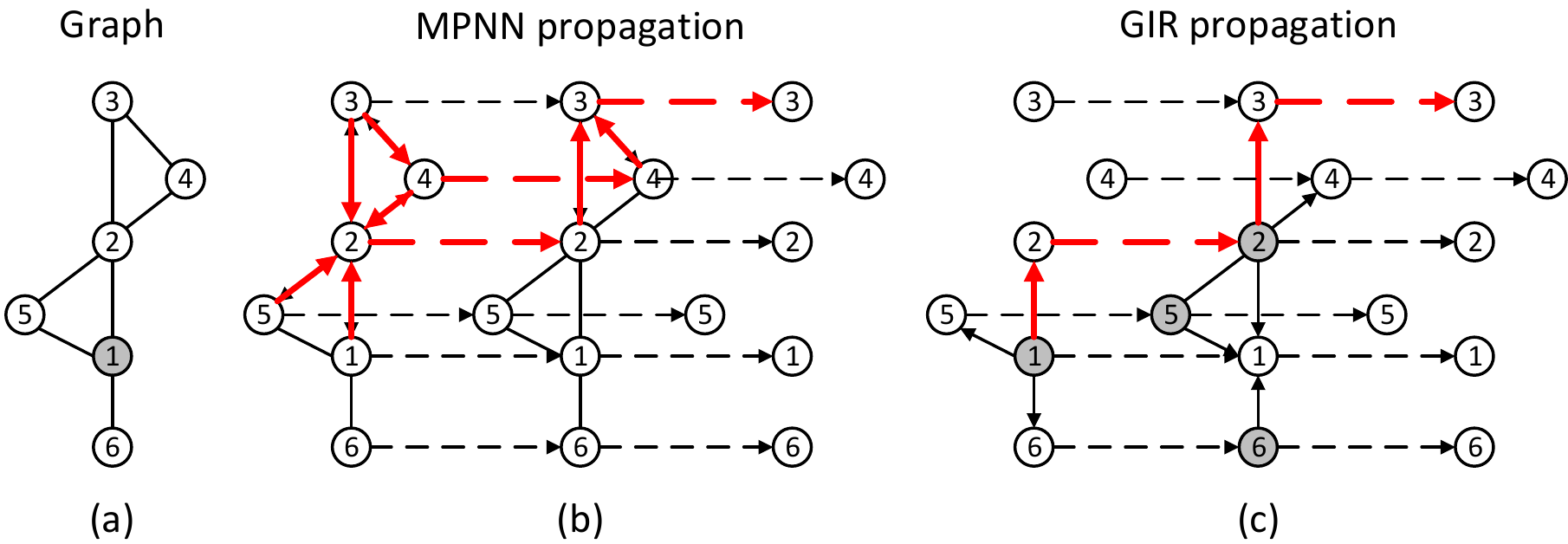}
  \caption{Message propagation paths for MPNNs and GIRs.
    \textbf{(a)}
    Example graph $\mathcal{G}$, nodes in grey are selected as anchors for anchor based GNNs.
    \textbf{(b\&c)}
    Message propagation paths on $\mathcal{G}$ for a two layer MPNN \textbf{(b)}
    and GIR \textbf{(c)}; solid line represents
    $Aggregate$; dot line represents $Update$.
    Grey nodes in \textbf{(c)} represent message passing sources on corresponding layer.
    Bold arrows in red highlight the propagation paths to node $v_3$.
    }
  \label{fig-gir}
\end{figure}

We propose Graph Inference Representation (GIR) model that
learns to integrate information from paths related to anchors for each node. 
GIRs propagate information on k-order ego graph of anchors,
which contains k-step paths starting from anchors.
For computing node representations,
GIRs apply MPNN layers step by step, propagate information from anchors to each
node (Figure \ref{fig-gir}, c).
Algorithm \ref{algo-gir} summarizes the framework of GIRs,
where remains some key components to be elaborated:
anchor selection and message passing functions.

\paragraph{Anchor Selection}

The intuition behind anchor based GNNs is that some nodes perform more important role
in the graph for the specific task \cite{liu2019gnn, Xu2020Graph}.
For getting position-aware embeddings on general tasks with no nodes
performing special role,
we heuristically prefer anchors to have high centrality and effective coverage
through k-step message passing.
We adopt GA-MPCA algorithm used in AGNN \cite{liu2019gnn} as default choice,
which select anchors according to degree centrality,
giving attention to coverage and information interaction intensity.  

\paragraph{Message Propagation}

GIRs use adapted MPNN layers for message propagation,
which contains neighborhood message passing and node state update steps
(Algorithm \ref{algo-gir}, line 6-7).
The key difference between GIRs and MPNNs is that the k-th layer of a GIR model only applies on
edges on k-step away from anchors instead of whole edges.
From the view of a specific node,
a k-layer MPNN model learns to encode its k-hop ego network information,
while a k-layer GIR model encode k-order representations related to anchors
(Figure \ref{fig-gir}).

Note that it is an expected manner for GIRs that
if the node $v$ has no propagation edge on the layer $l$,
its message in $l$ would be a zero vector,
and the MPNN layer $l$ would be degraded to a single layer perceptron for $v$.
Thus, GIRs can be viewed as a hybrid of MPNNs and MLPs.
From the view of GIRs, MPNNs use all nodes as anchors,
while MLPs use none.

We take the full-batch GraphSAGE layer \cite{hamilton2017inductive} as default
message propagation function choice,
concretely, we use mean function as aggregator and single layer perceptron as
update function.

\IncMargin{1.5em}
\begin{algorithm}
  \caption{The framework of GIRs}
  \label{algo-gir}
  \KwIn{Graph $\mathcal{G=(V,E)}$;number of layer $L$;node input attributes
    $\{x_v,\forall v\in \mathcal{V}\}$;anchor selection function $SelectAnchor$;neighborhood aggregator functions
    $Aggregate_l$;node update functions $Update_l$}
  \KwOut{A-position-aware embedding $\{z_v,\forall v\in \mathcal{V}\}$}
  $\mathcal{A}\leftarrow SelectAnchor(\mathcal{G})$\;
  $\mathcal{SRC} \leftarrow \mathcal{A}$ \tcc*{$SRC$ keeps the propagation source nodes}
  $h^0_v \leftarrow x_v, \forall v\in \mathcal{V}$\ \tcc*{$h_l^v$ represents
    hidden state of node $v$ on layer $l$}
  \For{$l\leftarrow 1$ \KwTo $L$}{
    \For{$v\in \mathcal{V}$}{
      $m_v \leftarrow Aggregate_l(\{h_u^{l-1},\forall u\in \mathcal{N}(v)\cap
      \mathcal{SRC}\})$ \tcc*{message on node $v$}
      $h_v^l\leftarrow Update_l(h_v^{l-1},m_v)$\;
    }
    $\mathcal{SRC} \leftarrow Sussessors(\mathcal{G},\mathcal{SRC})$\;
  }
  $z_v \leftarrow h_v^L$\;
\end{algorithm}
\DecMargin{1.5em}

\subsection{Position-aware GIRs}

In this part,
we show to what extent the GIR model helps break structural symmetry,
variants are proposed to get more powerful position aware ability in
the context of definition \ref{def-apos}.

\subsubsection{GIRs} 

Considering an unattributed graph $\mathcal{G}$,
for MPNNs, nodes with equivalent sub-structure will be indistinguishable.
GIRs truncate the message propagation path by anchors,
thus break sub-structural equivalence for nodes at different positions (Figure \ref{fig-pgnn}).

For commonly used order agnostic neighborhood aggregator (e.g.\ mean),
without properly identification,
every anchor node plays the same role,
thus position of each node related to a specific anchor node can not be
discriminated.
We show that the vanilla GIRs provide positioning ability with the overall
anchor node set.
We use commonly used mean function as aggregator,
appropriate designed MLPs as update functions to meet the assumption of
universal approximation theorem \cite{hornik1991approximation},
GIR propagation depth is assumed enough to cover all reachable nodes from anchors. 

\begin{theorem}
  \label{thm-gir}
  GIRs are \{$\mathcal{A}$\}-position-aware.
\end{theorem}

Theorem~\ref{thm-gir} states that for any graph there exists a parameter setting
of GIRs to get \{$\mathcal{A}$\}-position-aware embeddings,
while in practice, the learned model is dedicated to the downstream task.
The full proof of Theorem~\ref{thm-gir} is in the Appendix,
the proof construct a well-defined message propagate function (depicted in
Figure~\ref{fig-posgir}, a),
the existence of parameter setting is guaranteed by the universal approximation
theorem.

\subsubsection{GIR-MIX: multiple anchor set}

As GIRs are \{$\mathcal{A}$\}-position-aware,
if multiple anchor sets are further selected,
and applying GIRs on each,
more fine-grained positioning ability could be got.
We summarize the position-aware ability of this GIR-MIX model as follows:

\begin{corollary}
  GIR-MIXs with anchor sets
  $\mathnormal{A}=\{\mathcal{A}_1,\cdots,\mathcal{A}_n\}$
  are $\mathnormal{A}$-position-aware.
\end{corollary}

Instead of concatenating outputs only on the last layer,
we further extend the representation power by concatenate outputs in every
layer,
and send the fused hidden layer outputs to the next layer.
The GIR-MIX model could apply on transductive setting only,
for the order of anchor sets is aware.

\subsubsection{GIR-A: GIRs with anchor labeling}

Inspired by the one-hot labeling trick used in previous works,
and considering the concept of anchors in anchor based GNNs,
we extend node attributes by one-hot labeling for anchors,
and zero vector for other nodes.
It gives identification for anchors,
thus makes more accurate position.
Also, compared with global node one-hot labeling,
it has much lower input dimension.
As the anchor identification cannot be generalized,
the GIR-A model could apply on transductive setting only.

We show that with the anchor identification,
position related to a specific anchor could be represented.
Similar with the GIRs case,
we adopt mean aggregator,
update function model and propagation depth are properly chosen.

\begin{figure}
  \centering
  \includegraphics[width=0.75\linewidth]{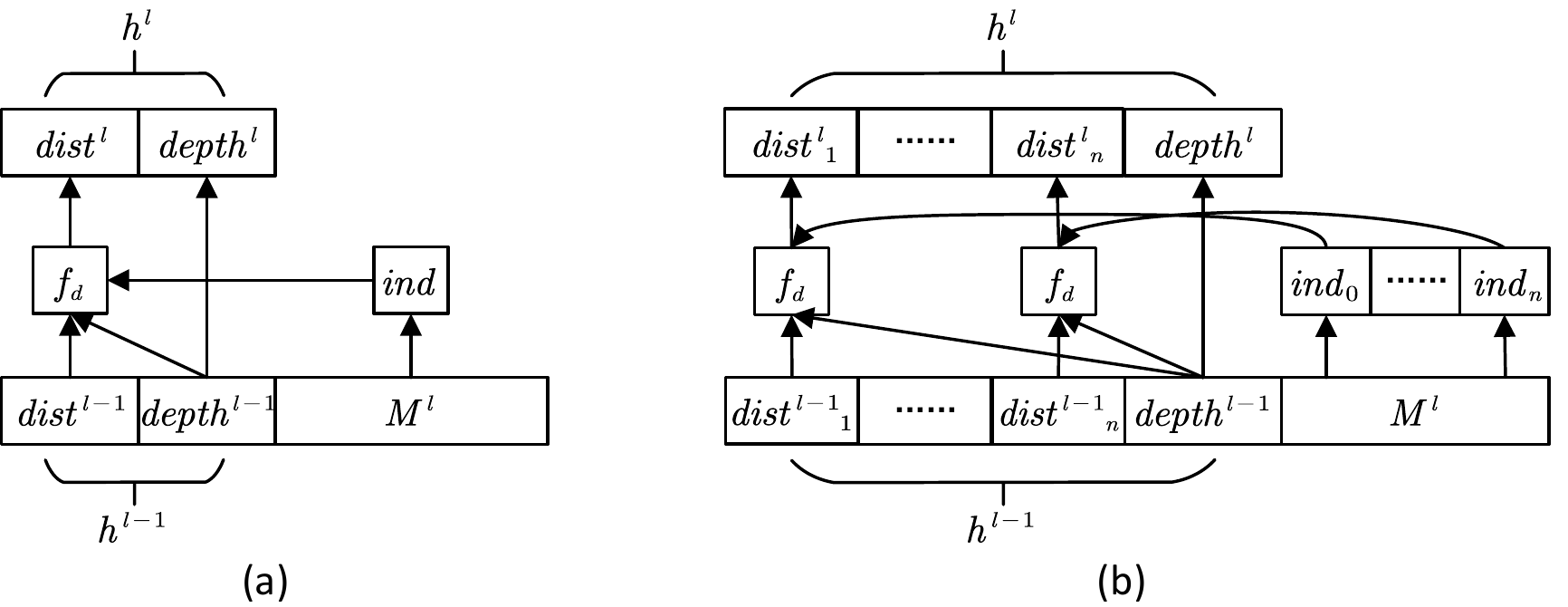}
  \caption{A construction of message passing function retaining distance information
    for GIRs \textbf{(a)} and
    GIR-As \textbf{(b)} }
  \label{fig-posgir}
\end{figure}

\begin{theorem}
  \label{thm-gira}
  GIR-As are $\mathcal{A}$-position-aware.
\end{theorem}

The full proof of Theorem~\ref{thm-gira} is in the Appendix,
the construction of the update function for the proof is depicted in Figure
\ref{fig-posgir}, b.

\subsection{GCN-GIR: multi view fusion}

\begin{figure}
  \centering
  \includegraphics[width=0.7\linewidth]{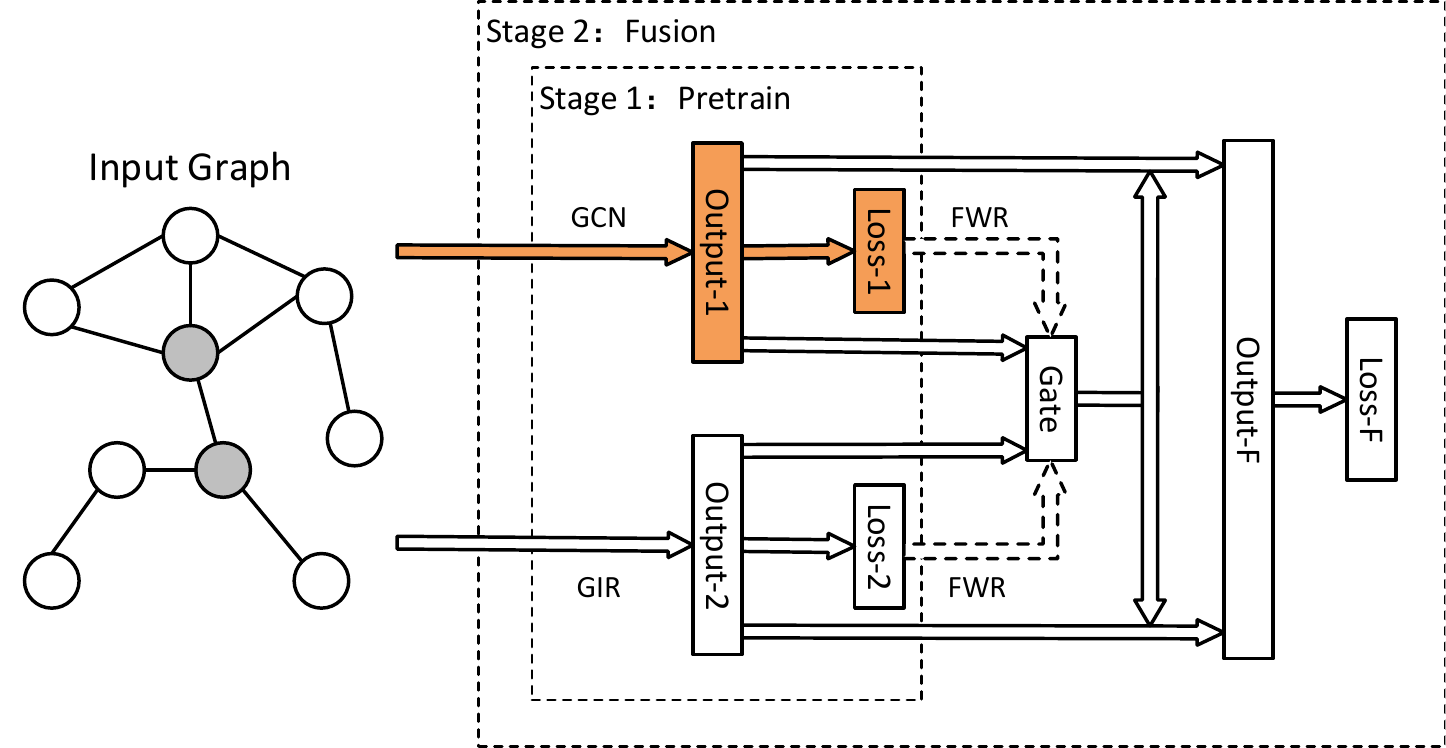}
  \caption{Overall architecture of GCN-GIRs.
    Stage 1 (pretrain stage) pretrain each expert alone,
    Stage 2 (fusion stage) freeze pretrained experts and train fusion part of
    the model.}
  \label{fig-fusion}
\end{figure}

One possible shortcoming for GIRs is that they work on only k-ego graph from anchors,
with other edges unrelated,
which may lose some useful information.
The influence of this issue depends largely on the choice of anchors.

In fact, considering that a GIR model encode information related to anchors for
each node,
it utilizes graph structure information in a biased interest,
thus learns another view of graph information compared with MPNNs.
Thus, it is a natural idea to exploit the possibility of fusing MPNNs and GIRs in
a complementary manner.
As an example model,
we take GCN and vanilla GIR as bases,
and terms it GCN-GIR.
Overall architecture of GCN-GIR is shown in Figure \ref{fig-fusion}.

We adopt decision fusion strategy,
and follow the idea of Mixture of Experts (MoE) \cite{jacobs1991adaptive}.
Some details would be focused
due to the non-independent nature of the graph data.
First, outputs of all experts are used instead of inputs to compute fusion gate.
This break the intuition of MoEs that dispatch independent data item to
appropriate classifier,
but more reasonable in this case as inputs of individual node may not enough
to make decision.
However, this strategy makes the gate for each expert relevant to other experts,
which may make training process unstable.
To alleviate this issue,
we adopt a two-stage training strategy,
which first pretrain each expert alone,
then freeze all experts, train only the fusion part of the model.

We further adopt fusion weight regularization (FWR)
\cite{shim2019robust},
which follow the idea that experts with lower loss matters more,
thus improve the consistency of the fusion gate and expert importances.

\section{Experiments}\label{section:experiments}

In this section,
we demonstrate performance of GIRs by two series of experiments.
We first take experiments on datasets from position-aware GNN model literature
to evaluate the performance of GIRs on position aware scenario.
To evaluate the representation bias of GIRs to MPNNs,
and the effect of multi view information fusion idea,
we take experiments on much larger scale node classification task from Open
Graph Benchmark \cite{hu2020ogb}.
All datasets are anonymous.

\subsection{Position-aware GIRs}

\paragraph{Datasets}

We take datasets from previous position-aware GNN literature
\cite{you2019position, li2020distance}, with MIT license.
Those datasets with too few scales or unfit for
transductive setting are filtered out,
for unattributed graphs,
we set node input attributes to ones as placeholders.
Details of the dataset are listed in Table~\ref{table-dataset}

\begin{table}
  \caption{Details of position-aware datasets.}
  \label{table-dataset}
  \centering
  \begin{tabular}{lccl}
    \toprule
    Dataset & $|\mathcal{V}|$ & $|\mathcal{E}|$ & Description \\
    \midrule
    Email & 920 & 14402 & a real-world communication graph from SNAP \cite{leskovec2007graph} \\
    Europe & 399 & 5995 & a flight network between airports in Europe \cite{ackland2005mapping} \\
    USA & 1190 & 13599 & a flight network between airports in USA \cite{ackland2005mapping} \\
    C.ele & 297 & 2148 & a neural network of C.elegans \cite{2006Nonoptimal} \\
    NS & 1461 & 2742 & a collaboration network between network science scientists \cite{newman2006finding} \\
    PB & 1222 & 16714 & a political post web-page reference network \cite{ackland2005mapping} \\
    
    \bottomrule
  \end{tabular}
\end{table}

\paragraph{Baselines}

We choose baselines from MPNNs and anchor based methods literature.
We choose representative method GCN \cite{kipf2017semi} as MPNN baseline,
for a fair comparison, random walk laplacian is used instead of symmetric
normalized laplacian
used in the origin paper.
PGNN and AGNN are chosen as anchor based baselines,
PGNN learns global position by random select anchors before each run,
while AGNN is a representative GNN method using fixed anchor.
Two variants of PGNN and AGNN are considered:
using truncated 2-hop shortest path distance (PGNN-E/AGNN-F)
and using exact shortest path distance (PGNN-E/AGNN-E).

We take experiments for vanilla GIR and several variants:
(1) GIR with multiple anchor sets (GIR-MIX);
(2) GIR with anchor labeling (GIR-A);
(3) GIR with node one-hot labeling (GIR-O).
As a general trick,
anchor labeling and node one-hot labeling are also applied in MPNN baselines (GCN-A/GCN-O).

\paragraph{Experimental setup}

Our experiments are taken on transductive setting.
For node pair level tasks,
80\% pairs are used for training,
and 10\% each as validation and test sets.
For node classification task,
60\% nodes are used for training,
and 20\% each as validation and test sets.
We report the test set performance at the best model with validation set,
and the final results are reported over 20 runs with different random seeds
on 5 random train/validation/test splits.

Our implementation is based on Deep Graph Library (DGL, Apache 2.0 licence) \cite{wang2019dgl},
with PyTorch (a BSD-style license) \cite{paszke2017automatic} backend,
the implementation of PGNN is adapted from the official implementation,
with MIT license.
Experiments are taken on single NVIDIA Tesla V100-SXM2-16GB GPU.
We use Adam optimizer for training,
with learning rate of 0.01, weight decay of 1e-5.
We use 3 layer models for all datasets;
hidden size and number of anchors are set heuristically according to the scale
of each dataset,
the number of anchors is also restricted by the GA-MPCA anchor selecting
algorithm.
For GIR-MIX, we simply divide anchors selected equally to form anchor sets,
we keep the number of anchor sets to be exactly divisible by hidden size and
the number of anchors,
and each anchor set contains more than one node.
Overall dataset aware hyperparameters are listed in Table~\ref{table-hp}.

\begin{table}
  \caption{Hyperparameters for experiments in position-aware scenarios.}
  \label{table-hp}
  \centering
  \begin{tabular}{lcccccc}
    \toprule
    Dataset & Hidden size & $|\mathcal{A}|$ & $|\mathnormal{A}|$ for GIR-MIX \\
    \midrule
    Email & 32 & 64 & 8 \\
    Europe & 16 & 8 & 4 \\
    USA & 32 & 64 & 8 \\
    C.ele & 16 & 16 & 8 \\
    NS & 32 & 64 & 8 \\
    PB & 32 & 64 & 8 \\
    \bottomrule
  \end{tabular}
\end{table}

\paragraph{Results and analysis}

\begin{table}
  \caption{Results on position aware scenario, measured in test ROC AUC (in \%) for link
    prediction (-lp) and node pair classification (-npc) task, and in test
    accuracy (in \%) for node classification (-nc) task.
  \textbf{Bolf font} highlights top-3 results, * highlights the best results.}
  \label{table-pos}
  \centering
  \begin{tabular}{lcccccc}
    \toprule
    & Email-npc & Europe-nc & USA-nc & C.ele-lp & NS-lp & PB-lp\\
    \midrule
    GCN & 51.2$\pm$1.4 & 21.7$\pm$2.5 & 23.7$\pm$2.3 & 48.7$\pm$3.8 & 55.0$\pm$5.7 & 34.4$\pm$4.1 \\
    GCN-A & 91.8$\pm$1.0 & 32.7$\pm$6.4 & 56.1$\pm$7.0 & 79.9$\pm$2.5 & 78.2$\pm$5.1 & 78.0$\pm$0.9 \\
    GCN-O & \textbf{97.4$\pm$0.4} & 32.1$\pm$5.3 & 52.4$\pm$4.1 & \textbf{84.3$\pm$2.2} & \textbf{94.5$\pm$0.6}* & 90.0$\pm$5.2 \\
    \midrule
    PGNN-F & 54.1$\pm$3.6 & \textbf{50.2$\pm$4.4} & \textbf{57.7$\pm$2.7} & 69.1$\pm$5.4 & 87.8$\pm$5.4 & 87.1$\pm$0.7 \\
    PGNN-E & 54.1$\pm$4.4 & \textbf{49.9$\pm$3.8} & 56.8$\pm$2.8 & 70.2$\pm$5.1 & \textbf{92.6$\pm$3.1} & 87.2$\pm$0.4 \\
    AGNN-F & 90.9$\pm$1.0 & 47.4$\pm$6.6 & \textbf{59.5$\pm$4.1} & 80.7$\pm$7.1 & 91.6$\pm$0.7 & 91.5$\pm$5.3 \\
    AGNN-E & 75.8$\pm$3.7 & 46.7$\pm$5.9 & 57.3$\pm$3.7 & 74.5$\pm$4.7 & \textbf{92.1$\pm$0.8} & 88.3$\pm$12.8 \\
    \midrule
    GIR & 50.2$\pm$0.7 & 30.8$\pm$4.5 & 29.9$\pm$2.6 & 55.5$\pm$2.7 & 78.6$\pm$4.7 & 59.7$\pm$4.1 \\
    GIR-A & \textbf{96.7$\pm$0.5} & \textbf{52.4$\pm$6.3}* & \textbf{60.4$\pm$4.7}* & \textbf{86.5$\pm$2.1} & 91.4$\pm$1.1 & \textbf{94.1$\pm$0.3} \\
    GIR-O & \textbf{99.1$\pm$0.2}* & 36.4$\pm$5.0 & 52.8$\pm$6.0 & \textbf{88.4$\pm$2.1}* & 89.6$\pm$2.6 & \textbf{94.4$\pm$0.3}* \\
    GIR-MIX & 82.9$\pm$3.7 & 44.1$\pm$5.1 & 56.7$\pm$4.5 & 78.8$\pm$8.3 & 90.5$\pm$1.3 & \textbf{92.0$\pm$0.4} \\
    \bottomrule
  \end{tabular}
\end{table}

Results of experiments are shown in table \ref{table-pos}.
With the same input feature,
GCN baseline fails to distinguish different nodes.
With $\{\mathcal{A}\}$-position-aware modeling ability,
vanilla GIR generally gets higher results,
yet inferior to more powerful position aware model structure.
More accurate positioning provided by GIR-MIX makes further improvement.
Benefiting by specially designed order-aware output layer and labeling trick,
PGNN and AGNN get relatively higher results.

Anchor labeling and node one-hot labeling improve performance compared to base
models by a large margin,
this demonstrates the position-aware modeling ability of those tricks.
Note that node one-hot labeling outperforms anchor labeling
in some datasets,
we owe this to its extra positioning ability than using anchors only.
This superiority will be degraded when more accurate positioning information matters.

GIR variants with those node identifier labeling tricks generally get outperformed results,
this demonstrates the power of our path information encoding structure.
The comparatively failed case on NS dataset is due to its sparsity,
unreachable nodes from anchors on the input graph will be embedded in to the
same representation due to the same input feature and the structure of GIR model.

\subsection{Multi view fusion}

\paragraph{Datasets and experimental setup}
We evaluate multi view fusion enhanced GIRs (GCN-GIRs) on ogbn-arxiv dataset from Open Graph
Benchmark (OGB) \cite{hu2020ogb}, with MIT license,
which is a citation network with 170k nodes and 1.2m edges.
Following standard experimental setup of OGB,
we use standard train/validation/test split
and report final test results over 10 runs.
The experiment is focusing on the effect of fusion strategy,
thus we simply take GCN as baseline,
and apply fusion strategy on GIR and GCN,
ablation study on fusion model are also taken.
Pretrained experts remain same for all two-stage model.

Experiments are taken on single NVIDIA Tesla V100-SXM2-16GB GPU.
We use LAMB optimizer \cite{You2020Large} for training,
with learning rate of 0.01, weight decay of 1e-5.
Number of hidden units is set to 256,
number of anchor nodes is set to 256.

\begin{table}
  \caption{Results for multi view fusion, measured in test accuracy (Acc, in \%).
    Expert complementarity is (EC, in \%) noted on
    fusion models.}
  \label{table-fusion}
  \centering
  \begin{tabular}{lccc}
    \toprule
    Model & Acc & EC & Notes \\
    \midrule
    GCN-3 & 72.22$\pm$0.14 & --- & 3-layer GCN\\
    GCN-5 & 72.29$\pm$0.19 & --- & 5-layer GCN  \\
    GIR-5 & 72.14$\pm$0.23 & --- & 5-layer GIR \\
    GCN-GCN & 72.52$\pm$0.12 & 6.60$\pm0.14$ & GCN-3 \& GCN-5 fusion\\
    \midrule
    GCN-GIR & \textbf{72.72$\pm$0.18} & 9.94$\pm$0.14 & GCN-3 \& GIR-5 fusion \\
    GCN-GIR-nf & 72.21$\pm$0.13 & 11.91$\pm$0.60 & no expert freeze on stage 2\\
    GCN-GIR-nFWR & 72.64$\pm$0.17 & 9.94$\pm$0.14 & no FWR on stage 2\\
    \midrule
    GCN-GIR-J & 71.70$\pm$0.19 & 16.58$\pm$0.45 & joint training with fused loss\\
    GCN-GIR-JA & 72.48$\pm$0.10 & 10.90$\pm$0.11 & use also uni-expert losses\\
    \bottomrule
  \end{tabular}
\end{table}

\paragraph{Measurements}

Apart from normally used test accuracy measurement,
we also note down the measurement indicating the complementarity.

We use uni-expert complementarity of i-th expert $EC_i$
to measure the difference between expert $i$ and other experts,
it is defined as the degree that other experts could correct the
decision of i-th expert,
formally,

\begin{equation}
  EC_i=harmonic\_mean(
  \frac{|\mathcal{S}^F_{i} \cap \mathcal{S}^T_{\sim i}|}{|\mathcal{S}^F_{i}|},
  \frac{|\mathcal{S}^F_{i} \cap \mathcal{S}^T_{\sim i}|}{|\mathcal{S}^T_{\sim i}|}) \\
\end{equation}

where $\mathcal{S}^{T/F}_{i/\sim i}$ represents the set of (truly/falsely)
classified nodes by
(i-th expert / one of experts except the i-th).
Overall expert complementarity (EC) measures the differences among experts from the
decision making perspective,
it is defined as the arithmetic mean of uni-expert complementarities.

\paragraph{Results and analysis}
Results of experiments are shown in table \ref{table-fusion}.
GCN-GIR shows much higher EC compared with GCN-GCN,
and the fusion strategy archives performance improvement for nearly all variants,
this support our statement that GIRs learn different information with typical
MPNNs,
and improvement could be achieved with proper fusion strategy.

Variants in the ablation study archive lower performance compared with GCN-GIR,
this demonstrates the effectiveness of our fusion strategy. 
Note that for those model without frozen pretrained experts,
higher ECs are got,
and joint training strategy without uni-expert learning objective (GCN-GIR-J)
gets the highest EC but lowest performance,
this support the need of expert-freeze strategy.
The FWR strategy also achieves improvement.

\section{Conclusion and future work}\label{section:conclusion}

We propose Graph Inference Representation (GIR),
a new type of anchor based Graph Neural Network (A-GNN),
together with its core variants,
for learning node embeddings incorporating node positional information using
path encoding.
Theoretical and experimental results show that GIRs benefit position-aware
graph learning.
Also, GIRs could improve performance on typical MPNNs by
providing complementary information with multi-modal fusion techniques.
For future work,
we suggest exploiting more on the anchor selecting strategy,
also, generalizing the defition of position-aware embeddings beyond shorest path distance
would be helpful.

\begin{ack}
\end{ack}

\small

\bibliography{graph_inference_representation_learning_graph_positional_embeddings_with_anchor_path_encoding}
\bibliographystyle{plainnat}

\appendix

\section{Appendix}

\subsection{Proof of Theorem 1}

\begin{proof}
  Given the basic structure and the design choices of GIRs,
  we restate theorem 1 as follows:
  
  There exists a function $g(\cdot)$ and properly parametered GIR model such
  that $g(GIR(\mathcal{G}, v_i))=d_{sp}(v_i,\mathcal{A})$, where
  $GIR(\mathcal{G}, v)$ represents the outputs on node $v$ of the parametered
  GIR model applied on graph $\mathcal{G}$.

  The propagation process of GIRs ensures that each node receives information
  from specific anchor node firstly along the shortest path between them,
  thus the shortest path distance of each node
  will be determined when message from anchors first reaches.
  We capture this observation in the distance update function,
  which keeps the current shortest path distance and the propagation depth, concretely,

  \begin{equation}
    \label{eq-gir:1}
    f_{d}(spd, depth, ind) =
    \left\{
      \begin{array}{ll}
        spd & ind=0 \\
        depth + 1 & ind=1, spd = 0 \\
        spd & ind=1, spd > 0 \\
      \end{array}
    \right.
  \end{equation}

  where $spd$ indicates the shorest path distance,
  $depth$ indicates current propagation depth,
  $ind$ indicates whether the message reaches.
  
  Then we construct the update function for l-th layer $f_l$ of the GIR model as
  follows:
  
  \begin{align}
    &h_v^l = f_l(h_v^{l-1}, M_{\mathcal{N}(v)\cap SRC_l}^{l})
      = \langle f_d(0,0,ind^l_v), 1 \rangle
            & l=1 \\
    &h_v^l = f_l(h_v^{l-1}, M_{\mathcal{N}(v)\cap SRC_l}^{l})
      = \langle f_{d}(spd_v^{l-1}, depth_v^{l-1}, ind_v^l), depth_v^{l-1}+1 \rangle
            &l>1
  \end{align}

  where $f_d$ is a function used to update the shortest path distance,
  $ind_v^l$ is the indicator of whether message from anchors on l-th layer reaches
  node $v$.
  On each layer $l$, for node $v$,
  if no neighbored source node exists,
  then $M_v^l$ would be a zero vector,
  so that the value of $ind_v^l$ could be determined,
  thus the $f_l$ is well-defined.
  The approximation of $f_l$ in GIRs is guarantied by universal approximation theorem.
  
  The output of the constructed $k$ layer GIR on node v becomes
  $z_v = h_v^k = \langle spd_v^k, depth_v^k \rangle$,
  then $(z_v)_0=d_{sp}(v,\mathcal{A})$,
  which concludes the proof.
  
\end{proof}

\subsection{Proof of Theorem 2}

\begin{proof}
  Similar to the proof of Theorem 1, we restate the theorem 2
  as:
  
  There exists series of functions $\{g_a(\cdot)|\forall a \in \mathcal{A}\}$
  and properly parametered GIR model such that $g_a(GIR(\mathcal{G},
  v))=d_{sp}(v,\{a\})$.

  Update function $f_l$ is constructed to keep the shortest path distance to
  every anchor node in $\mathcal{A}$ and current propagation depth,
  along with a vector indicating the k-order reachability from anchors.
  We apply distance update function \eqref{eq-gir:1} indepently for each anchor,
  and use reachability indicator to form $ind$, concretely,

  \begin{align}
    to\_ind(x) &= \left\{
      \begin{array}{ll}
        0 & x=0 \\
        1 & x>0
      \end{array}
    \right. \\
    ind_v^l \in \mathbb{R}^a &= 
    \left\{
      \begin{array}{ll}
        onehot(\mathcal{A}) & l = 1\\
        to\_ind(ind_v^{l-1}+mean(\{ind_u^{l-1}|u\in \mathcal{N}(v)\cap SRC_l\}))  & l>1 \\
      \end{array}
    \right.
  \end{align}

  where $onehot$ gives onehot encoding of $\mathcal{A}$.
  And the update function is constructed as follows:

  \begin{align}
    &\begin{array}{rl}h_v^l = & f_1(h_v^{l-1}, M_{\mathcal{N}(v)\cap SRC_l}^{l}) = \langle [ind^l_v]_1,\cdots,[ind^l_v]_{|\mathcal{A}|},1 \rangle
     \end{array}
    & l=1\\
    &\begin{array}{rll}
    h_v^l = & f_l(h_v^{l-1}, M_{\mathcal{N}(v)\cap SRC_l}^{l}) \\
    = & \langle f_{d}([spd_v^{l-1}]_1, depth_v^{l-1}, [ind_v^l]_1), \cdots,\\
            &f_{d}([spd_v^{l-1}]_{|\mathcal{A}|}, depth_v^{l-1}, [ind_v^l]_{|\mathcal{A}|}), \\
            &depth_v^{l-1}+1 \rangle
     \end{array}
    & l>1
  \end{align}

  The output of the constructed $k$ layer GIR on node $v$ becomes
  $z_v=h_v^k=\langle [spd_v^k]_1,\dots, [spd_v^k]_{|\mathcal{A}|},
  depth_v^k\rangle$,
  then $(z_v)_{id(a)}=d_{sp}(v, \{a\})$,
  where $id(a)$ represents the id of anchor ${a}$,
  which corresponds to the index of anchor onehot labeling.
  Here we have constructed the function $g(\cdot)$,
  thus concludes the proof.
\end{proof}

\end{document}